\documentclass[a4paper,11pt,twoside]{article}
\usepackage{graphicx}
\usepackage{booktabs}
\usepackage[latin1]{inputenc}
\usepackage[english]{babel}
\usepackage{amsfonts}
\usepackage[round,sort]{natbib} 
\usepackage{bm} 

\usepackage{caption}
\usepackage{array}
\usepackage{widetable}
\usepackage[T1]{fontenc}  
\usepackage{textcomp}  
\usepackage{txfonts}
\usepackage{mathptmx}
\usepackage{color}  
\usepackage{guit}   
\usepackage{url}
\usepackage{subfig}    
\usepackage{bm}

\makeatletter
\renewcommand\section{\@startsection {section}{1}{\z@}%
  {1\baselineskip \@plus .2\baselineskip \@minus .1\baselineskip}%
  {.5\baselineskip \@plus.1\baselineskip}%
  {\normalfont\bfseries}}

\renewcommand\subsection{\@startsection {subsection}{1}{\z@}%
  {1\baselineskip \@plus .1\baselineskip \@minus .05\baselineskip}%
  {.2\baselineskip \@plus .05\baselineskip}%
  {\normalfont\small\bfseries}}

\renewcommand\@seccntformat[1]{\csname the#1\endcsname.\hspace{0.2cm}}   
\makeatother

\begin{document}

\setcounter{page}{1} 

\begin{center}{\large
{\bf Multilingual textual data: an approach through multiple factor analysis  \vspace{1cm}}}
\end{center}
{\bf Belchin Kostov} \\
 Department of Statistics and Operational Research, Universitat Polit\`ecnica de Catalunya, C/ Jordi Girona 1-3, 08034 Barcelona, Spain
\\ \vspace{-3 mm}\\
{\bf  Ram\'on Alvarez-Esteban\footnote{Ram\'on Alvarez-Esteban, ramon.alvarez@unileon.es. ORCID 0000-0002-4751-2797}} \\
Department of Economics and Statistics, Universidad de Le\'on, Campus de Vegazana s/n, 24071 Le\'on, Spain
\\ \vspace{-3 mm}\\
{\bf  M\'onica B\'ecue-Bertaut} \\
 Department of Statistics and Operational Research,  Universitat Polit\`ecnica de Catalunya, C/ Jordi Girona 1-3, 08034 Barcelona, Spain
\\ \vspace{-3 mm}\\
{\bf Fran\c{c}ois Husson} \\
Institut Agro, Univ Rennes1, CNRS, IRMAR, 35000, Rennes, France
\\ \vspace{-3 mm}\\

\parindent 0pt
\begin{small} {\bf \textit{Abstract}} \textit{
This paper focuses on the analysis of open-ended questions answered in different languages.
Closed-ended questions, called contextual variables, are asked to all respondents in order to understand the relationships between the free and the closed responses among the different samples since the latter assumably affect the word choices.
We have developed "Multiple Factor Analysis on Generalized Aggregated Lexical Tables" (MFA-GALT) to jointly study the open-ended responses in different languages through the relationships between the choice of words and the variables that drive this choice. MFA-GALT studies if variability among words is structured in the same way by variability among variables, and inversely, from one sample to another.
An application on an international satisfaction survey shows the easy-to-interpret results that are proposed.
}

\vspace{6pt}
\end{small}

\vspace{8pt}
\baselineskip 14pt
\parindent 0.7cm

\section{INTRODUCTION}
\label{intro}
\noindent

Socio-economic surveys benefit from introducing open-ended questions in addition to closed-ended questions because they enrich each other. Closed-ended questions inform the interpretation of open-ended questions because the meaning of words is related to the characteristics or opinions of those who speak. For instance, in a satisfaction survey, customers are asked to rate certain aspects of the product and then freely give their opinion on which aspects could be improved, which is clearly linked to the ratings. In a survey that includes the question "What does health mean to you?" closed-ended questions such as gender, age, education and health status will greatly assist in exploring how the definitions of health vary with these variables.
In the case of international surveys, our framework, these open-ended questions raise the issue of analyzing the responses expressed in different languages by several samples.

In the case of a single language, textual statistics  \citep{Benzecri81,Lebart98} offer multidimensional tools for processing free responses. Separately for each sample, the free responses are encoded in the form of a frequency table respondents $\times$ words, called a lexical table (LT). A standard methodology is to apply correspondence analysis to this LT (CA-LT; direct analysis) and to use the closed information as a complement. It is also usual to group the responses of the categories of a closed question (such as age crossed with gender or education level, called contextual variable) and to build the frequency table of the words $\times$ categories, called aggregated lexical table (ALT) which can also be analyzed by CA (CA-ALT).

This approach is extended to several quantitative or qualitative contextual variables by using linearly constrained CA methods \citep{Takane91}. Thus, \cite{Balbi2001} address textual data including external information, \cite{Balbi2010} propose a double projection strategy by involving external information both on documents and words while \cite{Spano2012} apply canonical correspondence analysis (CCA; \cite{TerBraak86, TerBraak87}) to textual data. In line with these works,
\cite{Becue2014} and \cite{Becue2015} propose the method called CA on a generalized aggregated lexical table (CA-GALT). First, a table words $\times$ variables, called "Generalized Aggregated Lexical Table" (GALT), is developed by positioning the words on the variable-columns based on the values taken by the respondents who use them. Afterwards, this GALT is analyzed by means of a CCA adapted to textual data. In CA-GALT, like in any CA, both the variability of vocabulary through the variability of variables, and the variability of variables through the variability of vocabulary are explained. This fits perfectly with the perspective we have chosen to take here.

In the case of multilingual surveys, we propose to analyze simultaneously the different GALTs, one for each sample, by means of a multiple factor analysis (MFA; \citep{Escofier2016,Pages14}),
tailored to process a multiple GALT. This leads to the Multiple Factor Analysis for Generalized Aggregate Lexical Tables (MFA-GALT). This work outlines how to adapt MFA reasoning in order to handle a multiple GALT, and details its properties and graphical representations.

The aim of MFA-GALT is to jointly study the open-ended responses given by several samples in different languages through the relationships between the choice of words and the variables that motivate this choice. These relationships may or may not have similar structures. In other words, MFA-GALT studies whether variability among words is structured in the same way by variability among variables, and inversely, across samples.

The paper is organized as follows: Section \ref{sec:datastructure} presents the data structure and the notation. Section~\ref{sec:methods} recalls the principles of CA-GALT and MFA, the methods that form the basis of our approach; Section~\ref{sec:MFAGALT} is devoted to MFA adapted to multiple GALTs (MFA-GALT) and Section~\ref{properties} exposes the properties of the method. Enventually, the application of MFA-GALT to a full-scale application (Section~\ref{application}) shows its capabilities. The main conclusions are presented in Section~\ref{conclusion}.
\vspace{2pt}

\section{DATA STRUCTURE AND NOTATION}\label{sec:datastructure}
$L$ samples have answered a questionnaire with closed questions, either quantitative or categorical, all of identical type,
which constitute the contextual data. They have also answered an open-ended question in different languages conforming the textual data.
The $l$ sample has $I_l$ respondents who all together use $J_l$  different words in the $l$ language.
From these responses, we construct the ($I_l\times J_l$) table $\mathbf{Y_{l}}$, respondents $\times$ words; $N_l$ is the grand total for this table.

The responses to the closed-ended questions, common to all samples, are encoded in the $(I_l\times K)$ table $\mathbf{X_l}$, whose columns correspond either to quantitative variables or to dummy variables encoding the categories of one or more categorical variables. Whatever the type, $k$ and $K$ denote, respectively, the column variable $k$ and the total number of column variables. In what follows, the term \textit{variable} will be used for both types. From $\mathbf{Y_{l}}$, we compute the  ($I_l\times J_l$) proportion table $\mathbf{P_{l}}=\mathbf{Y_{l}}/N_l$.

If we consider only the sample $l$, the weights of the respondents are obtained from the margin of the rows of $\mathbf{P_{l}}$, thus proportional to the length of their free answers, and stored in the ($I_l\times I_l$)  diagonal matrix $\mathbf{D_{l}}$. The total weight of the respondents belonging to the same sample is equal to 1. In the same way, the weights of the words are obtained from the margin of the columns of $\mathbf{P_{l}}$, thus proportional to their counts, and stored in the  ($J_l\times J_l$) diagonal matrix $\mathbf{M_{l}}$. The total weight of the words used by the same sample is equal to 1. $\mathbf{X_{l}}$ is centered and possibly normalized in the case of quantitative variables, for the weighting system $\mathbf{D_{l}}$. The data structure including the relations between words and variables is the ($J_{l} \times K$) table $\mathbf{Q_{l}=\frac{Y_{l}^{T} X_{l}}{N_{l}}=P_{l}^{T} X_{l}}$. $\mathbf{Q_{l}}$ is called generalized aggregated lexical table.

\textbf{Remark}.
The name \textbf{Generalized Aggregated Lexical Table} and the acronym \textbf{GALT} are used to emphasize the great similarity between this table and the classic \textbf{aggregated lexical table} (\textbf{ALT}) developed in the case of a single categorical variable \citep{Lebart98}.

In fact, the calculation is exactly the same in both cases. What changes is only the expression of the matrix $\mathbf{X}$ itself. In the case of an ALT, this table is composed of the dummy variables corresponding to the categories of a single categorical variable.

\begin{figure}[!ht]
\center
\includegraphics[width=0.70\textwidth]{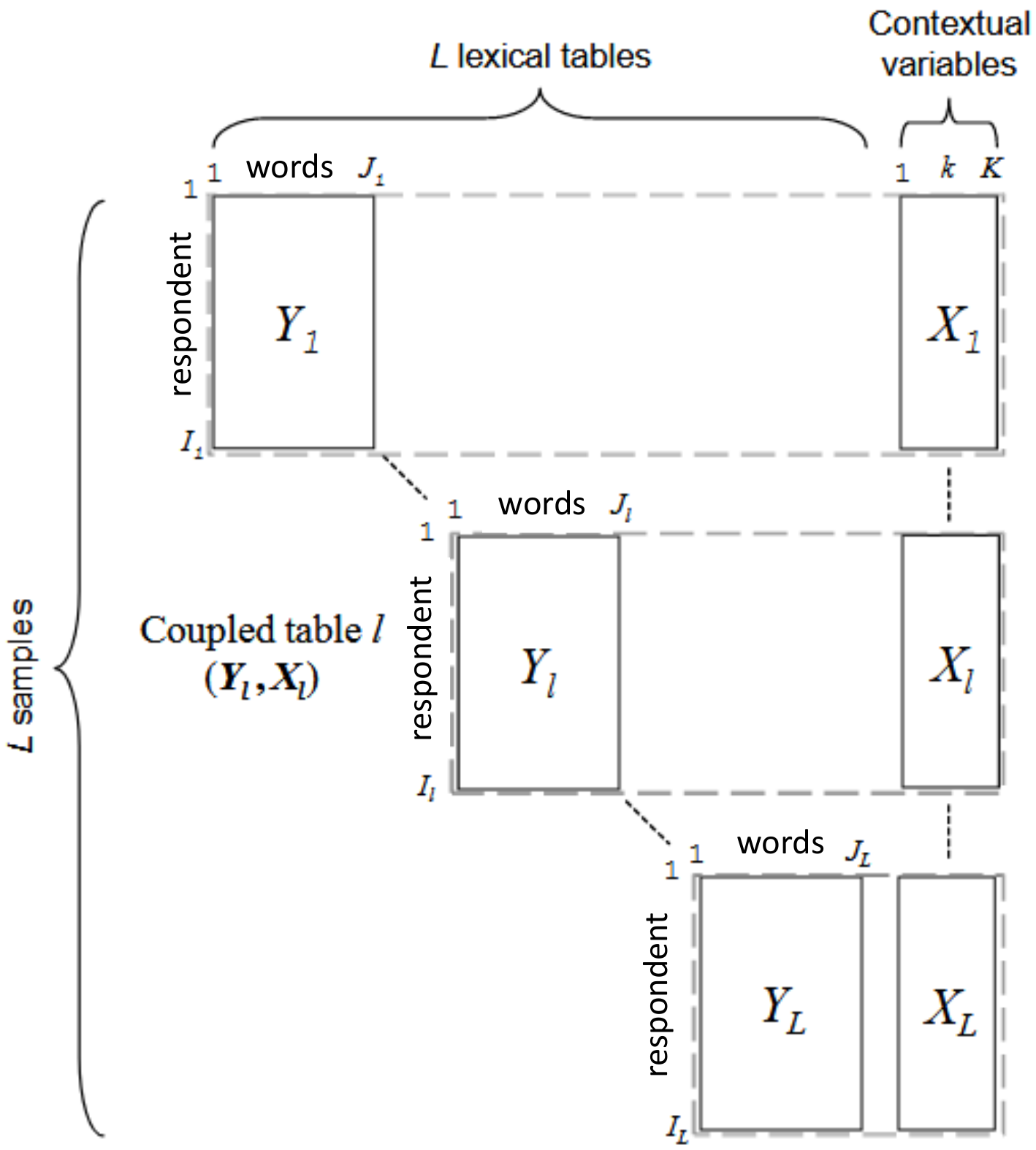}
\caption{Sequence of $L$ coupled tables}
\label{fig:structure}
\end{figure}

In the global analysis of the $L$ samples, we have to deal with $I=\sum_{l}I_{l}$ respondents who have used
$J=\sum_{l}J_{l}$ different words in the $N=\sum_{l}N_{l}$ occurrences that they pronounced through all the free answers.
 The respondent and word weights are resized so that both totals are equal to 1 for,
 respectively, the $I$ respondents and $J$ words. To this end, the respondent and word weights in sample $l$ are multiplied by $N_{l}/N$.
The respondent global weights are stored in the ($I \times I$)
 diagonal matrix $\mathbf{D}$. The word global weights are stored in the ($ J \times J$)
 diagonal matrix $\mathbf{M}$.
 \\
 The ($I \times K$) global table $\mathbf{X}$ is obtained by juxtaposing column-wise the $L$ tables $\mathbf{X_l}$,
 centered by set. As a consequence, table $\mathbf{X}$ is also centered for weighting system $\mathbf{D}$.

We suppose $K < J$. In the following, the symbols $I$, $I_l$, $J$, $J_l$, $K$, $L$ refer to both the set and its cardinal number.
\vspace{2pt}

\section{METHODS USED AS A BASIS OF OUR APPROACH}\label{sec:methods}

\subsection{DEALING WITH ONE SAMPLE}\label{subsec:Onesample}
In this section, we treat only one sample and therefore consider that it is not necessary to use the index $l$.

\subsubsection{CA-GALT method}\label{subsubsec:CAGALT-one}
We want to analyze the GALT $\mathbf{Q}$. To keep, as much as possible, a CA-like approach, we use CA-GALT method \citep{Becue2014,Becue2015} that we summarize hereafter.

Let the ($K \times K$) matrix $\mathbf{C=(X^{T} D X)}$ be the weighted correlation/covariance matrix of the variable-columns
of matrix $\mathbf{X}$. We calculate the ($J \times K$) matrix $\mathbf{Z}$, double standardized form of matrix
$\mathbf{Q}$:
\begin{equation} \mathbf{Z=M^{-1} Q C^{-1}}. \end{equation}
If $\mathbf{C}$ is not invertible, $\mathbf{C^{-1}}$ is substituted by the Moore-Penrose pseudoinverse $\mathbf{C^{-}}$.

Then, CA-GALT is performed through a principal component analysis (PCA)
in metrics $\mathbf{C}$ in the row space, and $\mathbf{M}$ in the column space, that is, PCA($\mathbf{Z,C,M}$).
This involves computing the $S$ ($S \leq K$) eigenvalues and eigenvectors of
\begin{equation}\label{eqdiag} \mathbf{Z^{T} M Z C}.\end{equation} The eigenvalues are stored
into the $(S\times S)$ diagonal matrix \boldmath$\Lambda$, and the eigenvectors into the $(K\times S)$ matrix \textbf{U}.
\\
CA-GALT is a dual projection analysis \citep{Becue2015} which leads to explain, on one hand, the variability of  words according to the variability of variables and, on the other hand, the variability of
variables according to the variability of words.

\textbf{Remark}.
Metric $\mathbf{C^{-1}}$ (or $\mathbf{C^{-}}$) operates a multivariate standardization that not only separately
standardizes the columns of $\mathbf{X}$, but in addition makes them uncorrelated \citep{brandimarte11, hardle12}.

\subsection{MFA GENERAL SCHEME}\label{subsec:MFA}

Multiple factor analysis \citep{Escofier2016, Pages14} analyzes multiple table, juxtaposing row-wise
either quantitative or categorical tables. It has been further extended to frequency tables \citep{Becue2004}.
 This method analyzes  a set of rows described by different sets of columns. The core of MFA is a PCA,
 with specific weights and metrics, applied to the multiple table dealing with the quantitative tables as
 in PCA, the categorical ones as in multiple correspondence analysis (MCA) and
 the frequency tables, in particular lexical tables, as in CA. The specific approach to each type of tables
  is obtained through a possible coding of the initial data and convenient choices of the weights and metrics.

In order to balance the influence of the sets on the first factorial dimension, the initial weights of the columns belonging to a given set are divided by the first eigenvalue resulting from the separate analysis of the corresponding table (PCA, MCA or CA depending on its type). Thus, the highest axial inertia of each set is normalized to 1.
MFA looks for identifying the main directions of variability in the data from a description of the rows by all the different sets of columns but balancing the importance of these sets.
MFA provides the classical results of principal component methods. PCA, MCA or CA characteristics and
interpretation rules are kept for, respectively, the quantitative sets, the categorical sets and the frequency sets.
MFA also offers graphical tools for comparing the different sets such as the superimposed global and partial representation
of the rows, as induced by all the sets or separately by each of them, as well as a synthetic representation of the sets,
in which each of them is represented by only one point. These graphical results allow us to compare the typologies
provided by each set in a common reference space.

\vspace{2pt}
\section{MFA ON MULTIPLE GALT}\label{sec:MFAGALT}
Hereafter, we tailor MFA to the case where the separate tables are GALT built from the different samples,
 that is to say, from the different coupled tables ($\mathbf{Y_{l},X_{l}}$) $(l=1,..,L)$. It is a question of inserting GALTs, and their analysis by CA-GALT, into this approach.

MFA, as described above, is usually devoted to a set of rows described by several sets of columns. Yet, we have
now to analyze several sets of row-words described by one set of column-variables. However, we are placed here
in a CA-like context where the role of rows and columns is exchangeable, which we could do without altering results.
In the following sections, we directly set out MFA-GALT method.

As classical MFA, MFA-GALT is performed in two steps.
First, each subtable - which is here a GALT - is analyzed separately, by applying the factorial method
corresponding to its type - here CA-GALT. In the second step, a global factorial analysis on all the sets
of multiple tables is performed, addressing each set as dealt with in the separate analyzes
but taking into account the reweighting used to balance the influence of the sets. In this way, the different
sets of rows have a similar influence on the first global axis. This reweighting consists in dividing the weights of
the rows of set $l$ by the first eigenvalue obtained in the separate analyzes of this set.
Therefore, the highest axial inertia of each set is standardized to 1. Among the properties of this
reweighting of the rows, note that the within-sets structures are not modified and,
except for very special cases, the first axis of the global analysis cannot be generated by a single table.
These two steps are detailed hereafter.
\\

\textbf{First step: separate analyzes}\\

Separate CA-GALTs are performed in each set, on the GALT $\mathbf{Q_{l}}$, following Section \ref{subsubsec:CAGALT-one}
scheme except for the metric used
in the row space (and weighting system in the column space). In this case, the covariance/correlation matrix computed
from all the respondents, that is, $\mathbf{C=(X^{T} D X)}$, is used, instead of matrices
$\mathbf{C_{l}=(X_{l}^{T} D_{l} X_{l})}$,
in all the separate analyzes.  This is due to the need to place all the row sets in a same metric space.
 In accordance with this, $\mathbf{C^{-1}}$ (or $\mathbf{C^{-}}$, if $\mathbf{C}$ is not invertible) is used
  to standardize $\mathbf{Q_{l}}$.
  So, in this first step $\mathbf{Z_{l}} = \mathbf{M_{l}^{-1} Q_{l} C^{-}}$ is analyzed
  via PCA($\mathbf{Z_{l}}$,$\mathbf{C}$,$\mathbf{M_{l}}$). The $L$ first eigenvalues $\lambda_1^{l}$
  will be used in the second step.
\\

\textbf{Second step: global analysis}\\

The row weighting system is updated to balance the influence of each set in the global analysis.
By construction, matrix $\mathbf{M}$ is divided into $L$ blocks. Block $l$ corresponds to the $J_{l}$ words used
in  sample $l$. The weights of the words of block $l$  are divided  by $\lambda_1^{l}$,
the first eigenvalue of the separate analysis of subtable $l$. The resulting weights are stored into the ($J\times J$) matrix
$\mathbf{M_{\lambda}}$.

The ($J\times K$) multiple table GALT $\mathbf{Q}$ juxtaposes column-wise the $L$ matrices
 $\mathbf{Q_{l}}$ but resized by multiplying them by coefficient $N_{l}/N$
 ($\mathbf{Q_{l}} \times N_{l}/N$). A double standardization of $\mathbf{Q}$, on the rows and  the columns,
  leads to the ($J\times K$) table $\mathbf{Z=M_{\lambda}^{-1} Q C^{-1}}$. If $\mathbf{C}$ is not invertible,
  $\mathbf{C^{-1}}$ is substituted by the by the Moore-Penrose pseudoinverse $\mathbf{C^{-}}$.
Then, \textsc{MFA-GALT} is performed through a non-standardized weighted PCA performed on the multiple table
\textbf{Z} with $\mathbf{M_{\lambda}}$ as row weights and  metric in the column space and $\mathbf{C}$ as
column weights and metric in the row space, that is, PCA($\mathbf{Z}$,$\mathbf{C}$,$\mathbf{M_{\lambda}}$).

\vspace{2pt}
\section{MAIN PROPERTIES OF MFA-GALT}\label{properties}

MFA-GALT provides the classical outputs of the principal components methods:
\begin{itemize}
\item coordinates, contributions and quality of representation of row-words
\item coordinates of categories at the centroid of the row-words used by respondents belonging to this category
\item coordinates of quantitative variables as covariances and/or correlation coefficients between factors and quantitative variables
\end{itemize}

Furthermore, outputs from MFA as a partial representation of the variables,
a synthetic representation of the sets and a measure of the similarity between the sets are also provided.

\subsection{REPRESENTATION OF THE ROW-WORDS AND THE COLUMN-VARIABLES}
PCA($\mathbf{Z}$,$\mathbf{C}$,$\mathbf{M_{\lambda}}$) involves diagonalizing
the matrix $\mathbf{Z^{T}M_{\lambda}ZC}$. The principal axis with rank $s$
corresponds to the eigenvector $\mathbf{u_s}$ ($\|\mathbf{u_s}\|_\mathbf{C}$=1)
associated with the eigenvalue $\lambda_{s}$:
\begin{equation} \label{diag1}\mathbf{Z^{T}M_{\lambda}ZC}\mathbf{u_s}=\lambda_{s}\mathbf{u_s}. \end{equation}
The eigenvalues $\lambda_{s}$ are stored into the ($S\times S$) diagonal matrix \boldmath$\Lambda$ and the eigenvectors $u_{s}$,
dispersion axes, into the columns of the ($K\times S$) matrix \textbf{U}.

By factor $s$  we mean the vector of coordinates on axis $s$ of either the word-rows (denoted $F_{s}$) or the
variable-columns (denoted $G_{s}$)  \citep{Benzecri73,Pages14}. The $S$ factors on the rows are stored into
the columns of the ($J\times S$) matrix \textbf{F}, calculated as:
\begin{equation} \mathbf{F=ZCU}. \end{equation}
The $S$ factors on the columns are stored into
the columns of the ($K\times S$) matrix \textbf{G}. Matrix \textbf{G} is calculated
through the use of the transition relations
between the factors on the rows and on the columns, also found in this case as in any PCA:
\begin{equation} \label{transG}\mathbf{G=Z^{T}M_{\lambda}F}\Lambda^{-1/2}.\end{equation}

\subsection{SUPERIMPOSED REPRESENTATION OF THE \textit{l} CLOUDS OF VARIABLES}
According to the $L$ sets of row-words, the column-variable $k$ of $\mathbf{Z}$ can be divided into $L$ subcolumns,
named partial variables and denoted $k^{l}$.
It is useful to represent simultaneously the $L$ partial scatter plots, each made up of the corresponding $K$ partial variables,
 on the same referential axes. To this end, we successively consider the $L$ matrices $\mathbf{Z_{l}}$,
 with dimension ($J_{l}\times K$), issued from matrix $\mathbf{Z}$ by keeping only the row-words belonging to set $l$.
 From these matrices, the ($J\times K$) matrices $\textbf{\~Z}_{l}$ are built by completing
 $\mathbf{Z_{l}}$ with 0 to be the dimension of $\textbf{Z}$.
In order to be represented on the global axes, the $K$ partial variables corresponding to the set $l$ are considered as
supplementary columns in the global analysis. Their coordinates are calculated using the transition relations
and stored in the ($K\times S$) matrix $\mathbf{G^{l}}$:

\begin{equation} \label{partial}\mathbf{G^{l}=}\textbf{\~Z}_{l}\mathbf{^{T}M_{\lambda}F}\Lambda^{-1/2}.\end{equation}

Therefore, the coordinates of the partial variables corresponding to set $l$ can be calculated
from the coordinates of the words used by only sample $l$. This relationship for partial variable $k^{l}$ is
expressed very simply
thanks to the structure of matrix $\textbf{\~Z}_{l}$ which contains only 0, except for the rows belonging to set $l$:
\begin{equation} \label{Gsk}
G_{s}(k^{l})=\frac{1}{\sqrt{\lambda_{s}}}\frac{1}{\sqrt{\lambda_1^{l}}}\sum_{j \in J_l}z_{jk}m_{jj}F_{s}(j).
 \end{equation}
In Eq.\ref{Gsk}, $[z_{jk}]$ denotes the generic term of $\textbf{Z}$  and $\frac{1}{\sqrt{\lambda_1^{l}}}m_{jj}$ denotes the generic term of
matrix $\mathbf{M_{\lambda}}$, being $m_{jj}$ the initial weight of the word $j$ (see Section
\ref{sec:datastructure}).

Accordingly to Eq.\ref{Gsk}, the partial variables relative to set $l$, are on the side of the words over-used in this sample
by respondents having high values for these variables.

Usually, all the "partial" variables can be represented on the same scatter plot,
informing about the similarities/dissimilarities among the samples.

\subsection{GLOBAL REPRESENTATION OF THE SETS}

Another result is to represent the $L$ groups on the same graph, each of them being represented by a point \citep{Pages14}.
To this end, the Lg coefficient, linkage measurement between one variable and one set of variables is used, applied here
to measure the linkage between each set and the axes that are retained. First, the ($K\times K$) matrix of scalar products
$\mathbf{W_l}$ between the $K$ column-variables of set $l$ is calculated as
\begin{equation} \mathbf{W_l=Z_l^{T}M_{\lambda l} Z_l}.\end{equation}
where the diagonal matrix $\mathbf{M_{l\lambda}}$, equal to block $l$ of matrix $\mathbf{M_{\lambda}}$, contains the weights
of the variables of set $l$, equal here to $\frac{1}{\lambda_{l}^{l}}$.
Then, $Lg(l,\mathbf{u_{s}})$ is calculated in the following way:
\begin{equation}Lg(l,\mathbf{u_{s}})=\langle \mathbf{W_{l}C,u_{s}C}\rangle=trace(\mathbf{W_{l}Cu_{s}u_{s}^{T}C}).\end{equation}

Further, $Lg(l,\mathbf{u_{s}})$ is used as a coordinate to place set $l$ upon axis of rank $s$. This coordinate always has a value
between 0 and 1. Accordingly, a map of all the sets, each one represented by one point, is obtained. This map also allows for visualizing
the similarity between the $L$ set structures.

\subsection{MEASURE OF THE ASSOCIATION BETWEEN VOCABULARY AND CONTEXTUAL VARIABLES}
Our proposal also includes the measurement of the association between vocabulary and contextual variables, first to select the variables that actually play a role and second to interpret the results. The measures, successively performed for each sample, are detailed in \cite{Becue2015}.

Briefly, vocabulary is said to be associated with a variable if the words differ significantly from each other in the values taken by the individuals using them. The association between a categorical variable and the vocabulary is evaluated with the classical chi-square test on the frequency table
crossing words and categories (=lexical table).

\begin{figure}
\centering
\includegraphics[width=0.85\textwidth]{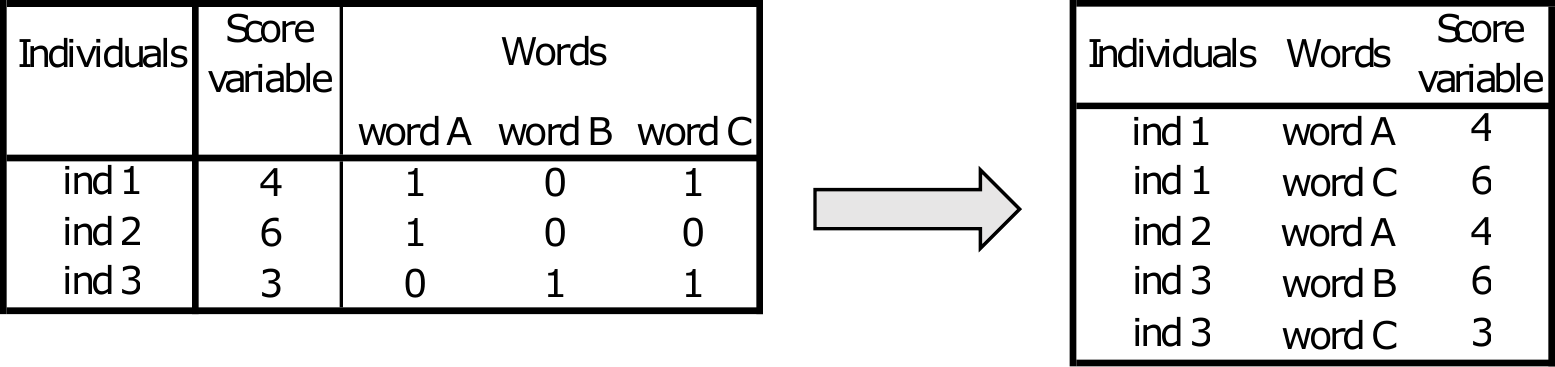}
\caption{Reorganization of the data for the one-way Anova that measures the association between vocabulary and a contextual variable.\label{fig:asso_data}}
\end{figure}

In the case of a quantitative variable, a one-way analysis of variance (Anova) is considered. The data table is reorganized as shown in Figure~\ref{fig:asso_data} before computing the one-way Anova: each row corresponds to an occurrence (i.e. a word cited by one individual). The score variable and the words variable have as many values as occurrences. Then the one-way Anova between the score and the words is performed, detecting whether relations between vocabulary and scores exist.

Note that, since the occurrences are not independent, the usual assumptions of Anova are not satisfied and it is better to use permutation tests.

\vspace{2pt}
\section{REAL DATA APPLICATION: INTERNATIONAL SURVEY}\label{application}
\begin{figure}
\centering
\includegraphics[width=0.8\textwidth]{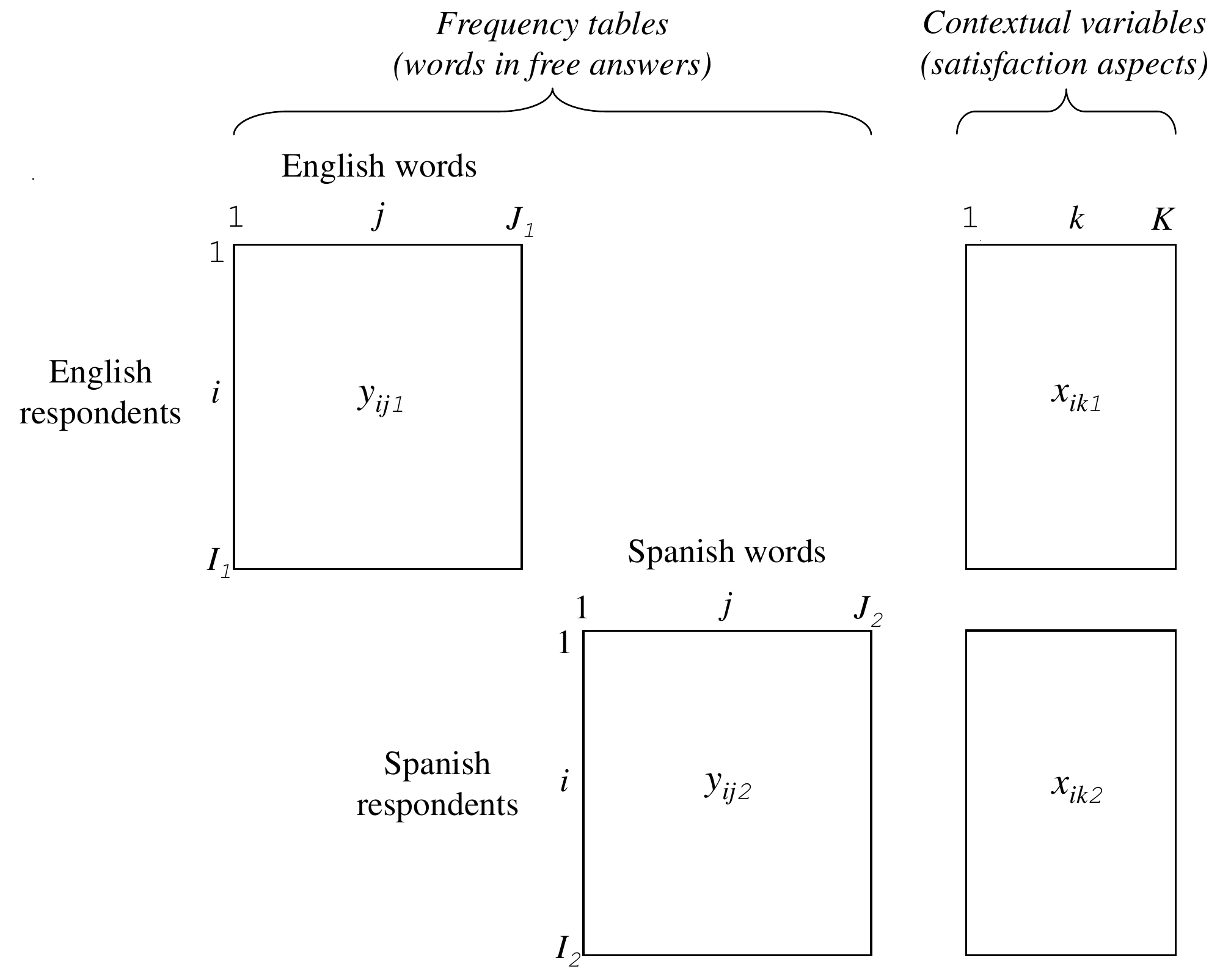}
\caption{The dataset. On the left, the lexical tables; on the right contextual variables.
In the example, $I_{1}$ = 283 (English respondents), $I_{2}$ = 977 (Spanish respondents), $J_{1}$ = 68 (English words),
 $J_{2}$ = 88 (Spanish words), K = 13 (satisfaction scores).\label{fig:example}}
\end{figure}

A railway company conducted a survey to find out the level of satisfaction of its passengers with the night trains that it offers.
Passengers were asked to rate their satisfaction about 13 aspects related to comfort
(general, cabin, bed, seat), cleanliness (common areas, cabin, toilet), staff attention (welcome attention, trip attention,
language skills) and others (cabin room, air conditioning, general aspects).
Each aspect was scored on a 11 point Likert scale, from 0 (very bad) to 10 (excellent).
In addition, an open-ended question was added asking for the aspects needing improvement.
This question required spontaneous answers which, in this case, were expressed in English or Spanish.
The data is stored into the data structure shown in Figure~\ref{fig:example}.

Preprocessing of the data includes a careful spelling correction for free responses. Stopwords are removed and then the words used at least 10 times are kept for the Spanish corpus (=all answers given in Spanish) whereas the threshold is 5 for the English corpus \citep{Lebart98, Murtagh05}. Finally, 977 respondents from the Spanish sample and 283 from the English one have no empty answers. Their response average is 3.1 long in both cases.
The Spanish corpus contains 3029 occurrences corresponding to 88 distinct words and the English corpus 871 occurrences corresponding to 68 distinct words.

For the score variables, the missing values were imputed. Note that for the graphics,  the grading
scale has been inverted. Thus, the largest values correspond to the greatest dissatisfaction, which makes the graphs easier to read.

\subsection{INITIAL FINDINGS}
The most frequent words give a first overview of the complaints, which are similar in both languages and stated with homologous words. \textit{Espacio/space}
 is too reduced, no place for \textit{maletas/luggages}. \textit{Cabinas/cabins} and \textit{asientos/seats} lack
 \textit{comodidad/comfort} while \textit{aseos/toilets} would benefit from more \textit{limpieza/cleanliness}.
 The \textit{Aire acondicionado/Air conditioning} seems to cause problems. For the English sample, the words \textit{staff} and \textit{English} are frequently cited. Aspects that were not asked to be evaluated are mentioned
 such as \textit{precio/price}.

 Table~\ref{table:contvariables} gives first insight with the means and standard deviations of the satisfaction scores.
\textit{Staff trip attention} obtains the highest score (8.07) for Spanish speakers while English speakers gave the highest score to \textit{Cabin cleanliness} (7.59). The lowest score is for \textit{Cabin room}
for both Spanish (5.33) and English speakers (5.71).
It is worth noting that the three aspects related to staff (\textit{Staff welcome}, \textit{Staff trip attention}
 and \textit{Staff language skills}) are clearly less appreciated by English speakers than for Spanish ones.

\begin{table*}
\caption{Mean satisfaction scores and association with vocabulary ratios\label{table:contvariables}}
\begin{center}
\fontsize{9}{8}\selectfont
    \begin{tabular}{ l  c  l  c  l}
    \hline\noalign{\smallskip}
    & \multicolumn{2}{c}{Spanish respondents} & \multicolumn{2}{c}{English respondents}\\
    Satisfaction scores & mean (SD) & ass.ratio (p-value) & mean (SD) &ass.ratio (p-value)\\
    \hline\noalign{\smallskip}
    General comfort & 6.82 (1.80) & 0.062 (<0.001) & 6.66 (1.95) & 0.091 (0.148) \\
    Cabin comfort & 6.37 (2.07) & 0.063 (<0.001) & 6.37 (2.03) & 0.121 (0.010) \\
    Cabin room & 5.33 (2.43 )& 0.089 (<0.001) & 5.71 (2.35) & 0.136 (<0.001) \\
    Bed comfort & 6.70 (1.98) & 0.050 (<0.001) & 6.72 (2.03) & 0.063 (0.918) \\
    Seat comfort & 6.10 (2.20) & 0.059 (<0.001) & 5.99 (2.38) & 0.123 (0.010) \\
    Air conditioning & 6.55 (2.55) & 0.107 (<0.001) & 6.51 (2.71) & 0.226 (<0.001) \\
    Common areas cleanliness & 7.41 (1.92) & 0.043 (<0.001) & 7.54 (1.86) & 0.082 (0.548) \\
    Cabin cleanliness & 7.59 (1.88) & 0.056 (<0.001) & 7.59 (1.81) & 0.116 (0.036) \\
    Toilet cleanliness & 6.21 (2.55) & 0.090 (<0.001) & 6.29 (2.40) & 0.150 (<0.001)\\
    Staff welcome attention & 7.99 (1.92) & 0.040 (0.018) & 7.29 (2.45) & 0.108 (0.062)\\
    Staff trip attention & 8.07 (1.85) & 0.038 (0.048) & 7.34 (2.29) & 0.092 (0.294)\\
    General aspects & 7.77 (1.65) & 0.038 (0.034) & 7.48 (1.91) & 0.079 (0.590)\\
    Staff language skills & 7.72 (2.08) & 0.052 (<0.001) & 7.14 (2.52) & 0.154 (<0.001)\\
    \hline\noalign{\smallskip}
    \end{tabular}
\end{center}
\end{table*}

The association between vocabulary and a contextual variable (see Table~\ref{table:contvariables}, columns \emph{ass.ratio (p-value)}) shows that \textit{Air conditioning} obtains the highest ratio for both Spanish (0.107) and English (0.226) speakers. \textit{Toilet cleanliness} ranks second for this indicator for Spanish speakers (0.090) while
\textit{Staff language skill} is placed second with 0.154 in the case of English ones, although \textit{Toilet cleanliness}, with 0.150 is
the third and not far away. Note that \textit{Staff language skill} is only ranked eighth for this indicator in the case of Spanish speakers. \textit{Cabin room} is ranked the third for Spanish speakers and the fourth for English speakers.

\subsection{MFA-GALT ON THE MULTILINGUAL DATASET}

Ranking aspects from the association-with-vocabulary ratio does not coincide with the score-average ranking. This implies that, according to passengers' opinions, the aspects that should mostly be improved do not correspond to the aspects they are less satisfied with. This justifies the interest in collecting information through open-ended questions, as this is different and complementary information.

MFA-GALT is applied on the multiple generalized aggregated lexical table. The total inertia is equal to 9.91.
The first eigenvalue (1.75 which corresponds to 17.70\% of the total inertia) is close to the number of sets, which means that the two sets share the dispersion direction corresponding to the first global axis. The second eigenvalue (1.42, 14.36\% of the total inertia) and the third  (1.23, 12.39\% of the total inertia) are close but the following eigenvalues are much smaller, which leads us to only focus on the first three axes.
To avoid overemphasizing the example, we will only interpret the first 2 axes. For a more detailed description of the results, and in particular of the 3rd dimension, the reader can refer to the thesis of \cite{kostov:pca}.

\begin{figure*}[!ht]
    \centering
    \subfloat[Best represented satisfaction scores\label{fig:scores12}]{%
      \includegraphics[width=0.49\textwidth]{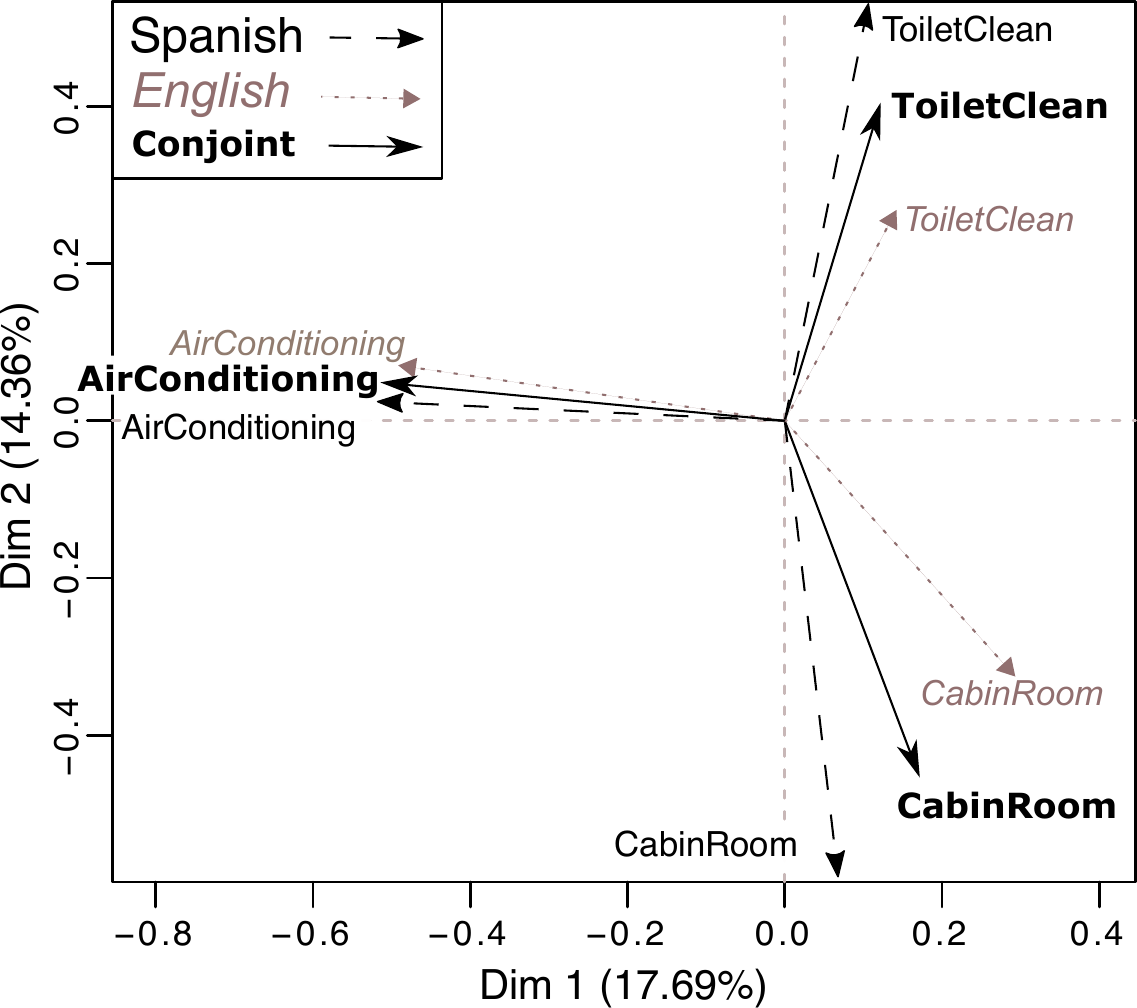}
    }
    \hfill
    \subfloat[Most contributive words\label{fig:words12}]{%
      \includegraphics[width=0.49\textwidth]{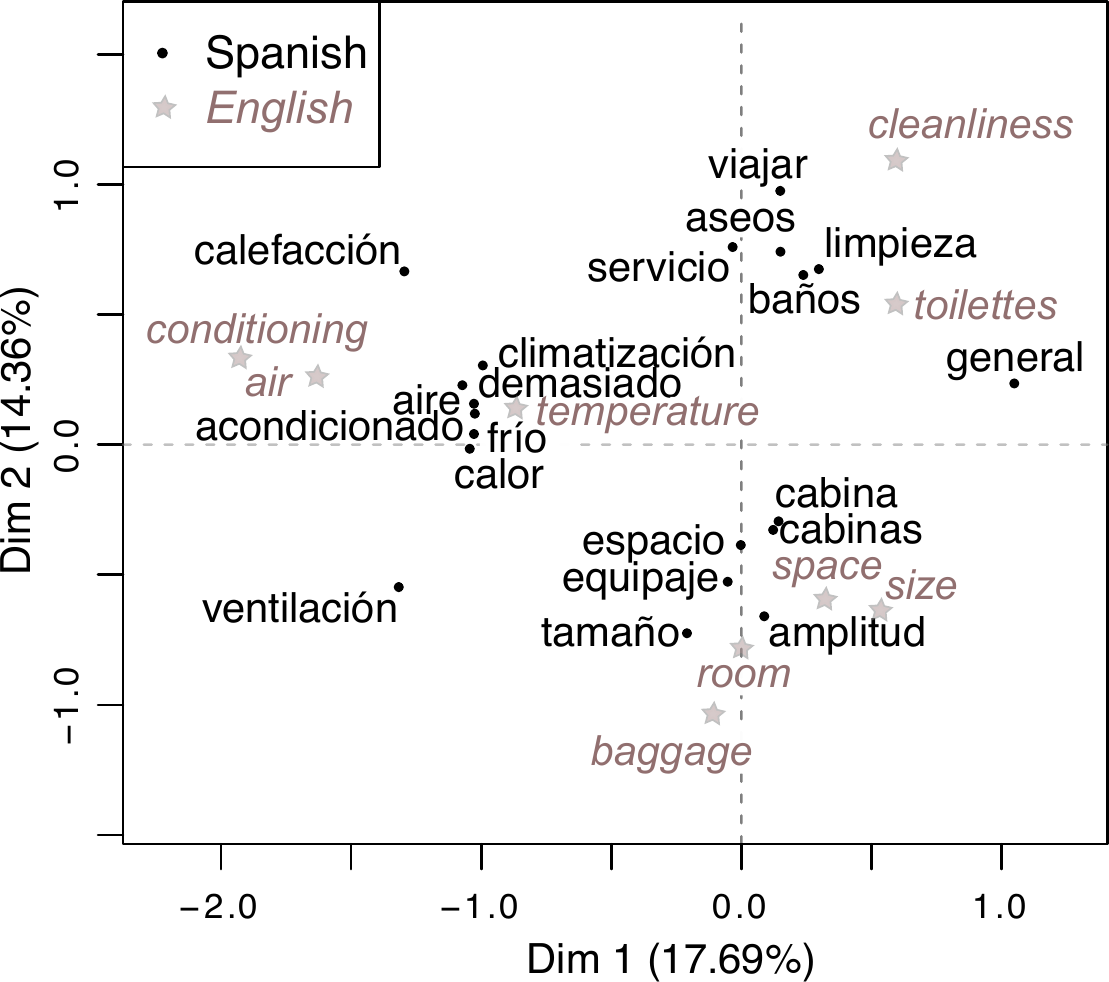}
    }
    \caption{MFA-GALT: Representation of variables (i.e. scores) and words on the plane (1,2)\label{fig:scoreswords}}
  \end{figure*}

\subsubsection{Global representation of the satisfaction scores and words}
MFA-GALT provides graphical results in which each variable (each score) points to the words associated with it. It thus indicates the defects of the scored aspect, either shared or not in both languages. Figure~\ref{fig:scores12} shows the best represented satisfaction scores on the first MFA-GALT principal plane through their covariances with the axes. To avoid overloading the graphics, only the scores that are well represented are shown (here, those that have a square cosine summed on the two axes over 0.5).
 We first only look at the global representations of the scores which offers a three polar structure. The three poles refer to inconveniences associated with \textit{Air conditioning}, lack of \textit{Toilet cleanliness} and problems related to \textit{Cabin room}.
This corresponds well with what the association-with-vocabulary ratios suggested.
Figure~\ref{fig:words12} represent the words, Spanish and English, that contribute more than twice to the contribution average. We can then see words highly associated with \textit{air conditioning}, showing its drawbacks: \textit{air}/\textit{aire}, \textit{conditioning}/\textit{acondicionado}, \textit{temperature}, \textit{fr\'{\i}o} (=cold)  \textit{climatizaci\'{o}n} (=air conditioner), \textit{ventilaci\'{o}n}/\textit{ventilation} and \textit{calefacci\'{o}n} (=heating). On the positive part of the second axis, the lack of \textit{Toilet cleanliness} is characterized by \textit{cleanliness}/\textit{limpieza}, \textit{toilettes}/\textit{aseos}/\textit{ba\~{n}os}. On the negative part, the problems with \textit{Cabin room} are described with the words \textit{size}/\textit{espacio} and \textit{cabins}/\textit{cabina}/\textit{cabinas}.

In this example, axis 3 is specific to one set, the English one, that has problems with the staff speaking poor English.

\begin{figure*}[!ht]
    \centering
      \includegraphics[width=0.49\textwidth]{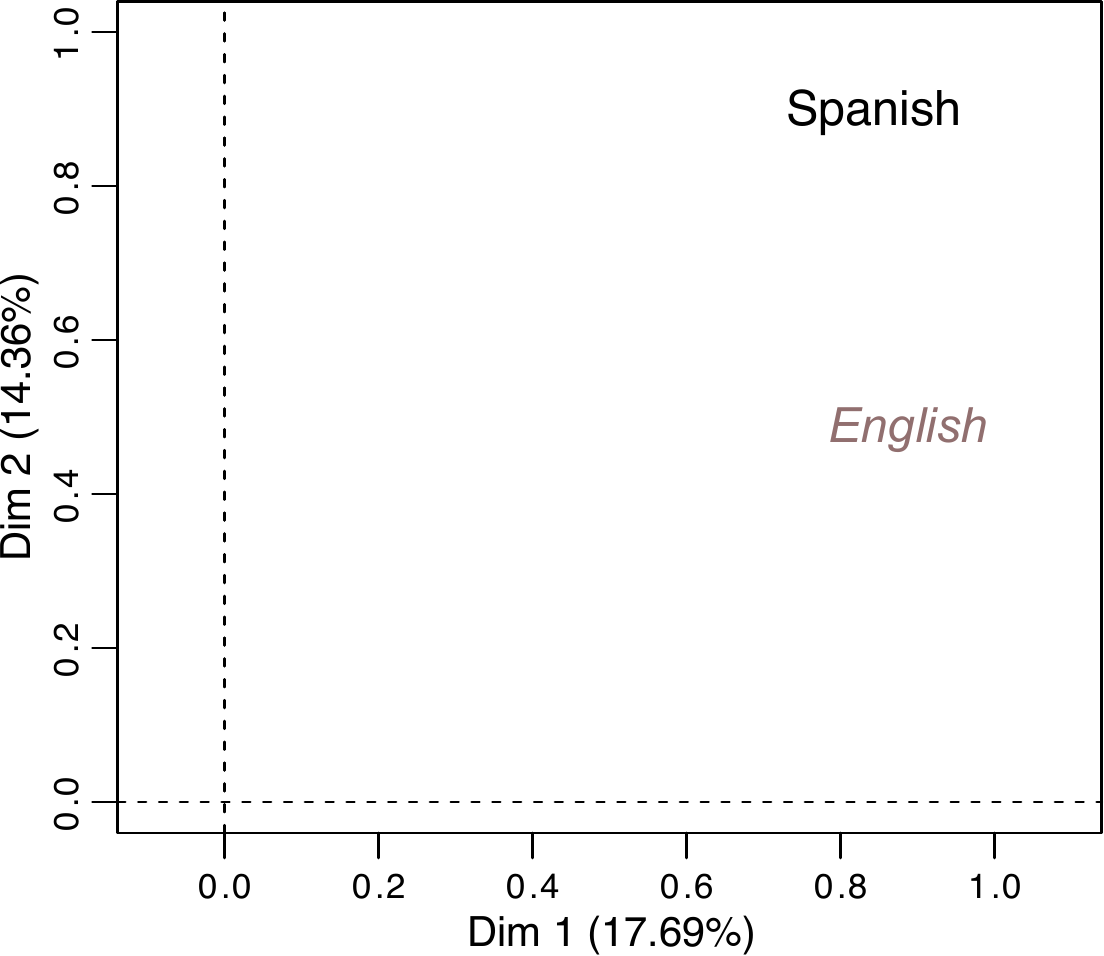}
    \caption{Representation of the sets.\label{fig:groups}}
  \end{figure*}

\subsubsection{Partial representation of the satisfaction scores}
Figure~\ref{fig:scores12} shows the superimposed representation of the global and partial representations of the satisfaction scores on the plane (1,2). It highlights the similarities and differences between the two sets regarding the association between words and scores.
\textit{Air conditioning} has a similar behavior in both sets on the first axis. On the second axis, \textit{Toilet cleanliness} and \textit{Cabin room} are more strongly associated with Spanish vocabulary than with English vocabulary, which translates into higher covariances; the complaints of the former seem more accentuated, giving rise to more words. On the third dimension, only English speakers complain about the lack of \textit{staff language skills}.

\subsubsection{Representation of the sets}
Similarity measures confirm that both sets share dispersion directions. The value of the RV coefficient, multivariate generalization of the squared Pearson correlation coefficient,  equal to 0.74 (p<0.001) confirms that the partial configurations are relatively close but not homothetic.

According to the representation of the sets on the first dimension, the coordinate of Spanish sample is 0.85
whereas the coordinate of English sample takes a slightly higher value (0.91) (Figure~\ref{fig:groups}).
It means that the first axis provided by MFA-GALT is an axis of high importance for both sets. Consequently this is a common dispersion structure. On the other hand, the Spanish set has a much larger (0.91 vs. 0.51) coordinate on the second axis. This means that the MFA-GALT second axis is of high importance for Spanish speakers, not so much for English speakers. The opposite is observed for the third axis (0.42 for Spanish vs. 0.81 for English).

\vspace{2pt}
\section{CONCLUSION}\label{conclusion}
This paper proposes an original principal component method to deal with open-ended questions answered in different languages. This type of textual and contextual data leads to a sequence of coupled tables,
each one made of one frequency table (=lexical table) and one quantitative/qualitative table.
We tackle these data through the relations between the words and the contextual variables.
Two methods are combined, \textsc{CA-GALT} and \textsc{MFA}, hence the name of the new method: \textit{Multiple Factor Analysis on Generalized Aggregated Lexical Tables} (\textsc{MFA-GALT}). The first one places the words of the different sets in a same space generated by the variables, which results in the construction of the GALTs. The second one enables the simultaneous analysis of these tables in such a way that the MFA properties are preserved.

The international survey with open questions answered in different languages was analyzed with \textsc{MFA-GALT}. This made it possible to study similarities among words from the same language, similarities among homologous words from different languages, associations between words and satisfaction scores, similarities between satisfaction score structures (partial representations) and similarities between groups. The results of this application show that \textsc{MFA-GALT} provides a good synthesis of the data through easy-to-interpret outputs.

The R package \texttt{Xplortext} includes the function \texttt{LexGalt} which allows the implementation of the CA-GALT and MFA-GALT methods.

\addcontentsline{toc}{section}{\refname}
\bibliographystyle{sa-ijas}  
\bibliography{biblio}   

\end{document}